\documentclass[10pt,twocolumn,letterpaper]{article}

\pdfoutput=1
\usepackage{cvpr}
\usepackage{times}
\usepackage{epsfig}
\usepackage{graphicx}
\usepackage{amsmath}
\usepackage{amssymb}

\usepackage{algorithm}
\usepackage{algorithmic}

\cvprfinalcopy 

\def\cvprPaperID{685} 

\ifcvprfinal\fi
\begin{document}

\title{Approximate Nearest Neighbor Fields in Video}

\author{Nir Ben-Zrihem\\
Department of Electrical Engineering\\
Technion, Israel\\
{\tt\small bentzinir@gmail.com}
\and
Lihi Zelnik-Manor\\
Department of Electrical Engineering\\
Technion, Israel\\
{\tt\small lihi@ee.technion.ac.il}
}

\maketitle


\begin{abstract}
%
We introduce RIANN (Ring Intersection Approximate Nearest Neighbor search), an algorithm for matching patches of a video to a set of reference patches in real-time.
For each query, RIANN finds potential matches by intersecting rings around key points in appearance space.
Its search complexity is reversely correlated to the amount of temporal change, making it a good fit for videos, where typically most patches change slowly with time.
Experiments show that RIANN is up to two orders of magnitude faster than previous ANN methods, and is the only solution that operates in real-time.
We further demonstrate how RIANN can be used for real-time video processing and provide examples for a range of real-time video applications, including colorization, denoising, and several artistic effects.
\end{abstract}

\thispagestyle{empty}

\section{Introduction}
\label{sec:intro}

The Approximate Nearest Neighbor (ANN) problem could be defined as follows: given a set of reference points and incoming query points, quickly report the reference point closest to each query.
Approximate solutions perform the task fast, but do not guarantee the exact nearest neighbor will be found. 
The common solutions are based on data structures that enable fast search such as random projections~\cite{LSH,ann-1}, or kd-trees~\cite{k-dTree,FLANN}.

When the set of queries consists of all patches of an image the result is an ANN-Field (ANNF). Barnes et al.~\cite{PatchMatch} developed the PatchMatch approach for ANNF, which shows that spatial coherency can be harnessed to obtain fast computation. Algorithms that integrate traditional ANN methods with PatchMatch' spatial coherency yield even faster runtimes~\cite{HeSun,CSH}. It is thus not surprising that patch-matching methods have become very popular and now lie in the heart of many computer vision applications, e.g., texture synthesis~\cite{texturesynthesis}, image denoising~\cite{patch-Match-denoising}, super-resolution~\cite{example-based-SR}, and image editing~\cite{PatchMatch,general-PatchMatch,Image-Summarization} to name a few.

\begin{figure}
  \centering
  \includegraphics[width=0.4\textwidth]{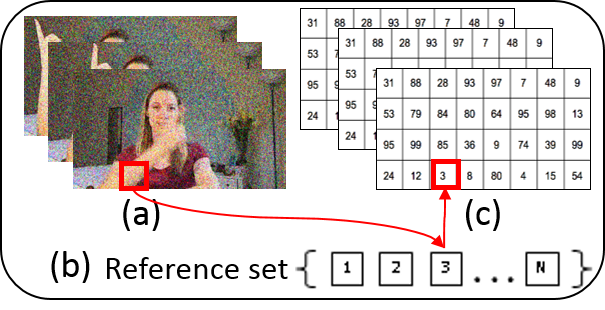}\\
  \caption{Video ANN Fields: (a) Input: Live stream of video frames. (b) A reference set of image patches. (c) For each frame, a dense ANN Field is produced by matching patches from the reference set. This is done in realtime.}\label{fig:videoANNF}
\end{figure}

In this paper we present an efficient ANNF algorithm for video.
The problem setup we address is matching all patches of a video to a set of reference patches, in real-time, as illustrated in Figure~\ref{fig:videoANNF}.
In our formulation the reference set is fixed for the entire video.
In fact, we could use the same reference set for different videos.
Additionally, our reference set of patches is not restricted to originate from a single image or a video frame, as is commonly assumed for image PatchMatching.
Instead, it consists of a non-ordered collection of patches, e.g., a dictionary~\cite{ksvd,regression-tree}.
This setup enables real-time computation of the ANN Fields for video.
We show empirically, that it does not harm accuracy.

Our algorithm is designed to achieve low run-time by relying on two key ideas.
First, we leverage the fact that the reference set is fixed, and hence moderate pre-processing is acceptable. At pre-processing we construct a data structure of the reference set, that enables efficient indexing during run-time.
%
Second, we rely on temporal-coherency to adapt our hashing functions to the data.
The hashing we use is not fixed a-priori as in previous works~\cite{PatchMatch,CSH,TreeCANN,HeSun}, but rather it
is tuned per query patch during runtime.
In regions with high temporal change the hashing is tuned to be coarse, which leads to higher computation times (larger bins translate to checking out more candidate matches).
On the contrary, in regions with low change, the hashing is finer and hence the computation time is lower.
We refer to this approach as ``query-sensitive hashing''.
Our hashing is generic to any distance metric.
We show examples for working with 2D spatial patches, however, our algorithm is easily extended to work with 3D spatio-temporal patches as well.

To confirm the usefulness of the proposed approach we further discuss how it can be adopted for real-time video processing.
We show that a broad range of patch-based image transformations can be approximated using our nearest neighbor matching.
Specifically we provide examples of denoising, colorization and several styling effects.

The rest of this paper is organized as follows.
We start by reviewing relevant previous work on image and video ANN in Section~\ref{sec:related-work}.
We then present our hashing technique and its integration into a video-ANN Fields solution in Section~\ref{sec:QShashing}.
We compare the performance of our algorithm to previous work in Section~\ref{sec:evaluate} and suggest several applications in Section~\ref{sec:ann-based-video-proessing}.
Further discussion is provided in Section~\ref{sec:coherency-spatial} and our conclusions are laid out in Section~\ref{sec:conclusion}.

\section{Related Work}
\label{sec:related-work}

The general problem of Approximate Nearest Neighbor matching received several excellent solutions that have become highly popular~\cite{LSH,ann-1,ann-2,k-dTree,TSVQ}.
None of these, however, reach real-time computation of ANN Fields in video.
Image-specific methods for computing the ANN Field between a pair of images achieve shorter run-times by further exploiting properties of natural images~\cite{PatchMatch,general-PatchMatch,CSH,HeSun,TreeCANN,FLANN}.
In particular, they rely on spatial coherency in images to propagate good matches between neighboring patches in the image plane.
While sufficiently fast for most interactive image-editing applications, these methods are far from running at conventional video frame rates.
It is only fair to say that these methods were not designed for video and do not leverage statistical properties of video.

An extension from images to video was proposed by Liu \& Freeman~\cite{liu2010highquality} for the propose of video denoising through non-local-means. For each patch in the video they search for $k$ Approximate Nearest Neighbors within the same frame or in nearby frames. This is done by propagating candidate matches both temporally, using optical flow, and spatially in a similar manner to PatchMatch~\cite{PatchMatch}.
One can think of this problem setup as similar to ours, but with a varying reference set. While we keep a fixed reference set for the entire video,~\cite{liu2010highquality} use as different set of reference patches for each video frame.

Another group of works that find matches between patches in video are those that estimate the optical flow field~\cite{zach2007duality,sun2010secrets,horn1981determining,brox2004high}. Several of these achieve real-time performance, often via GPU implementation. However, the problem definition for optical flow is different from the one we pose. The optical flow field aims to capture the motion in a video sequence. As such, matches are computed only between consecutive pairs of frames. In addition, small displacements and smooth motion are usually assumed.
Solutions for large-displacement optical-flow have also been proposed~\cite{brox2011large,xu2012motion,chen2013large,sundaram2010dense}.
The methods of~\cite{brox2011large,xu2012motion,sundaram2010dense} integrate keypoint detection and matching into the optical flow estimation to address large displacements.
Chen et al.~\cite{chen2013large} initiate the flow estimation with ANN fields obtained by CSH~\cite{CSH}.
None of these methods are near real-time performance, with the fastest algorithms running at a few seconds per frame.
Furthermore, while allowing for large displacements, their goal is still computing a smooth motion field, while ours is obtaining similarity based matches.

In Section~\ref{sec:ann-based-video-proessing} we show how our ANNF framework can be used for video processing.
The idea of using ANNF for video processing has been proposed before, and several works make use of it.
Sun \& Liu~\cite{sun2012non} suggest an approach to video deblocking that considers both optical flow estimation and ANNs found using a kd-tree.
An approach that utilizes temporal propagation for video super-resolution is proposed in~\cite{liu2011bayesian}. They as well rely on optical flow estimation for this.
The quality of the results obtained by these methods is high but this comes at the price of very long runtimes, often in the hours.

Our work is also somewhat related to methods for high dimensional filtering, which can be run on patches in video for computing non-local-means.
Brox et al.~\cite{brox2008efficient} speed-up traditional non-local-means by clustering the patches in a tree structure.
Adams et al.~\cite{adams2009gaussian} and Gastal et al.~\cite{gastal2012adaptive} propose efficient solutions for high-dimensional filtering based on Gaussian kd-trees.
None of these methods provide real-time performance when the filter is non-local-means on patches (unless harsh dimensionality reduction is applied).

\section{Ring Intersection Hashing}
\label{sec:QShashing}

Current solutions to computing the ANN Field between a pair of images utilize two mechanisms to find candidate matches for each query patch:  spatial coherency, and appearance based indexing. Some algorithms rely only on the first, while others integrate both.
To allow for a generic reference set (rather than an image) we avoid reliance on spatial coherency altogether.
Instead, we rely on an important characteristic of video sequences: there is a strong temporal coherence between consecutive video frames. We harness this for efficient appearance-based indexing.

\subsection{Temporal Coherency}

\begin{figure}[t]
  \centering
  \begin{tabular}{cc}
  \includegraphics[width=0.2\textwidth]{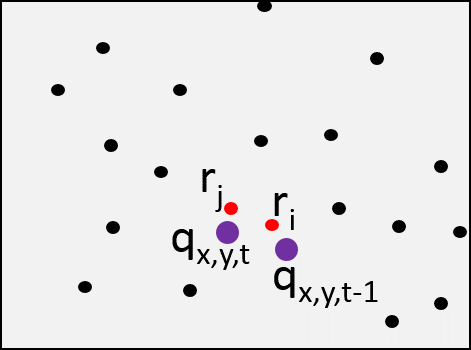}&
  \includegraphics[width=0.2\textwidth]{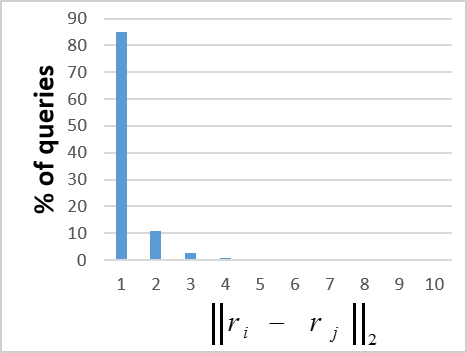}\\
  (a) & (b)  
  \end{tabular}
  \caption{Coherency in appearance:
  (a) When consecutive query patches $q_{x,y,t-1}$ and $q_{x,y,t}$ are similar, their exact NNs in the reference set $r_i$ and $r_j$ are also similar.
  (b) The histogram depicts the distribution of distances between such pairs of reference patches $r_i$ and $r_j$. It shows that for most queries $q_{x,y,t}$, the exact NN lies in a $d=1$ radius ball around the NN of the predecessor patch $q_{x,y,t-1}$.}\label{fig:appearanceNgbors}
\end{figure}

Let $q_{x,y,t}$ denote the query patch $q$ at position $x,y$ in frame $t$ of the input video.
Our first observation is that there is strong temporal coherency between consecutive patches $q_{x,y,t-1},q_{x,y,t}$. In other words, most patches change slowly with time, and hence patch $q_{x,y,t-1}$ is often similar to patch $q_{x,y,t}$. Our second observation is that temporal coherency implies coherency in appearance space: if patch $q_{x,y,t-1}$ was matched to patch $r_i$ of the reference set, then patch $q_{x,y,t}$ is highly likely to be matched to patches similar to $r_i$ in appearance. This is illustrated visually in Figure~\ref{fig:appearanceNgbors}.(a).

To evaluate the strength of coherency in appearance we performed the following experiment. For each $8\times8$ patch in a given target video we find its exact nearest neighbor in a reference set of patches. For this experiment the reference set consists of all patches of a random frame of the video. Let patches $r_i,r_j$ be the exact NNs of patches $q_{x,y,t-1}$ and $q_{x,y,t}$, respectively. We then compute the distance in appearance between the matches: $d=\|r_i-r_j\|_2$.
This was repeated over pairs of consecutive patches from 20 randomly chosen videos from the Hollywood2~\cite{hollywood2} data-set.
The distribution of distances $d$ is presented in Figure~\ref{fig:appearanceNgbors}.(b), after excluding patches where $r_i=r_j$ (effectively excluding static background patches). As can be seen, for $\sim85\%$ of the patches $d\leq1$. This implies that coherency in appearance is a strong cue that can be utilized for video ANN.

\subsection{Adaptive Hashing}

These observations imply that candidate matches for patch $q_{x,y,t}$ should be searched near $r_i$, the match of $q_{x,y,t-1}$. Furthermore, the search region should be adapted per query. In areas of low temporal change, a local search near $r_i$ should suffice, whereas, in areas of high temporal change the search should consider a broader range.

\begin{figure*}[t]
  \centering
  \begin{tabular}{cccc}
   \includegraphics[width=0.21\textwidth]{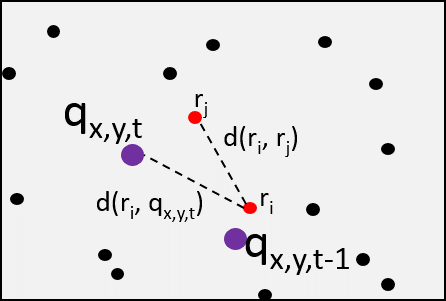}&
   \includegraphics[width=0.21\textwidth]{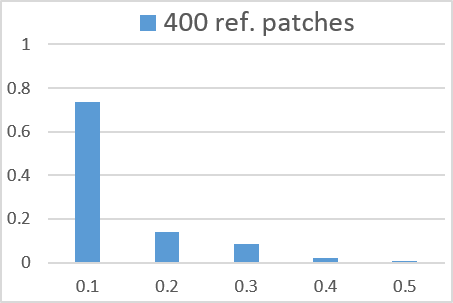}&
   \includegraphics[width=0.21\textwidth]{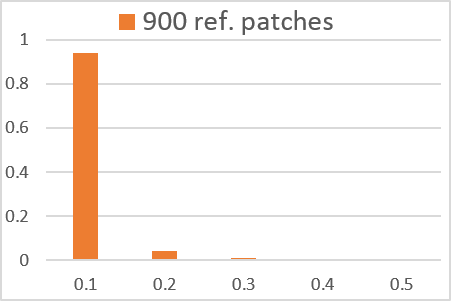}&
    \includegraphics[width=0.21\textwidth]{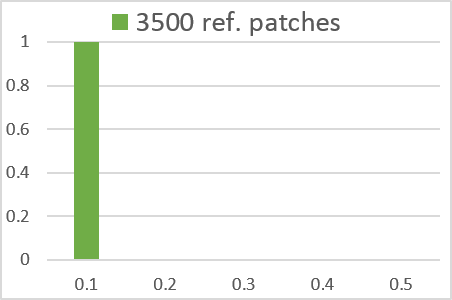}\\
    \end{tabular}
   $|dist(r_i,r_j)-dist(r_i,q_{x,y,t})|$
  \caption{Predicting search radius:
  (Left) The distance $dist(r_i,q_{x,y,t})$ is a good predictor for the distance $dist(r_i,r_j)$.
  (Right)
  Histograms of $|dist(r_i,r_j)-dist(r_i,q_{x,y,t})|$ for three sizes of reference sets. This suggests that $dist(r_i,q_{x,y,t})$ is a strong predictor for $dist(r_i,r_j)$. The correlation becomes stronger as we use a larger reference set. To exclude static background from our analysis we include only queries where $r_i\ne r_j$. Statistics were computed over 20 videos from the Hollywood2 database.}
  \label{fig:predictor}
\end{figure*}

So, how do we determine the search area for each query?
The answer requires one further observation.
If the reference set is adequate then $q_{x,y,t}$ and its NN $r_j$ are very close to each other in appearance space. This suggests that the distance between $r_i$ and $r_j$ is close to the distance between $r_i$ and $q_{x,y,t}$, i.e., $dist(r_i,r_j)\approx dist(r_i,q_{x,y,t})$.
This is illustrated and verified empirically in Figure~\ref{fig:predictor}, for three sizes of reference sets.
As can be seen, for set size of $900$ patches or more, this assertion holds with very high probability.

\begin{figure}[tb]
  \centering
  \includegraphics[scale=0.32]{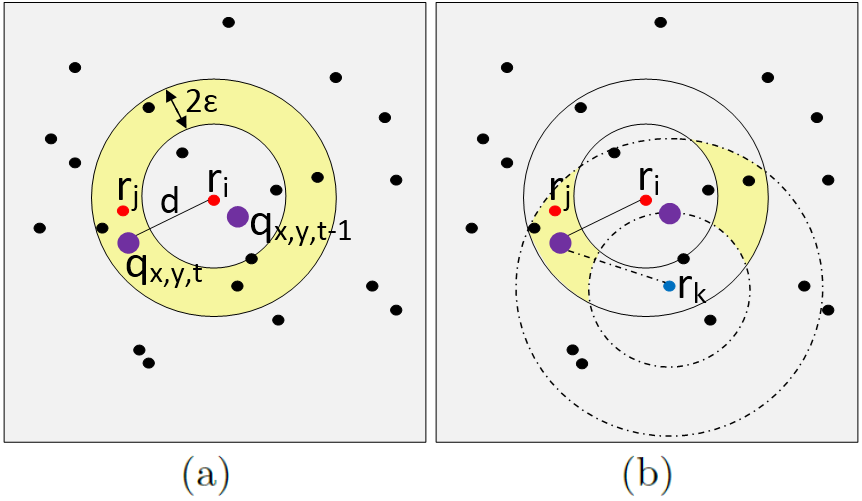}\\
  \caption{ Ring Intersection Hashing:
  (a) To find candidate neighbors for $q_{x,y,t}$ we draw a ring of radius $d=dist(r_i,q_{x,y,t})$ around $r_i$. Here $r_i$ is the match found for $q_{x,y,t-1}$.
  (b) To exclude candidates that are far from $q_{x,y,t}$, we draw another ring, this time around $r_k$, one of the current candidates. We continue to add rings, and leave in the candidate set only those in the intersection.}\label{image-ringHashing}
\end{figure}

Based on the above, to find the match $r_j$ we should search through a ``fat'' ring of radius $\approx dist(r_i,q_{x,y,t})$, around $r_i$, as illustrated in Figure~\ref{image-ringHashing}.(a). In areas with significant change $q_{x,y,t}$ will be far from $r_i$, the ring radius will be large and will include many reference patches. On the contrary, in areas with little change $q_{x,y,t}$ will be near $r_i$, the ring radius will be small and will include only a few candidate reference patches. The width of the ring, denoted by $2\varepsilon$, could also be tuned to adapt the search space.
We take the width $\varepsilon$ to be proportional to the radius $\varepsilon= \alpha \cdot dist(r_i,q_{x,y,t})$. In our implementation and all experiments $\alpha=0.25$. Our rings are thus wider in areas of large changes and narrower in regions of little change.


As can be seen in Figure~\ref{image-ringHashing}.(a), the ring around $r_i$ includes the neighbors of $q_{x,y,t}$, but it also includes reference patches that are very far from $q_{x,y,t}$, e.g., on the other side of the ring. To exclude these patches from the set of candidates we employ further constraints. Note that our observations regarding $r_i$ are true also for any other point in the reference set. That is, if $dist(r_j,q_{x,y,t})$ is small, then $dist(r_k,r_j)\approx dist(r_k,q_{x,y,t})$ for any patch $r_k$ of the reference set. Therefore, we further draw rings of radius $dist(r_k,q_{x,y,t})\pm\varepsilon$, around points $r_k$ selected at random from the current set of candidates. The final candidate NNs are those that lie in the intersection of all rings, as illustrated in Figure~\ref{image-ringHashing}.(b). For each query we continue to add rings until the number of candidate NNs is below a given threshold $L$. In all our experiments the threshold is fixed to $L=20$ candidates. The final match is the one closest to $q_{x,y,t}$ among the candidate NNs.

\subsection{Computational Complexity}

As a last step of our construction, we need to make sure that the rings and their intersections can be computed efficiently during runtime. Therefore, at the pre-processing stage we compute for each reference patch a sorted list of its distances from all other reference patches. The computation time of this list and the space it takes are both $O(n^2)$, where $n$ is the size of the reference set.

Having these sorted lists available, significantly speeds up the computation at runtime. For each query patch $q_{x,y,t}$, we compute the distance $d=dist(q_{x,y,t},r_i)$, where $r_i$ is the last known match. We add all reference points of distance $d\pm\varepsilon$ from $r_i$ to the set of candidate NN. Thus, the computation complexity for each ring includes one distance calculation, and two binary searches in a sorted array, $O(\log{n})$. We continue to add rings and compute the intersection between them, until the number of candidates $\le L$.

\begin{figure}[htb]
  \centering
  \begin{tabular}{c}
  \includegraphics[width=0.65\linewidth]{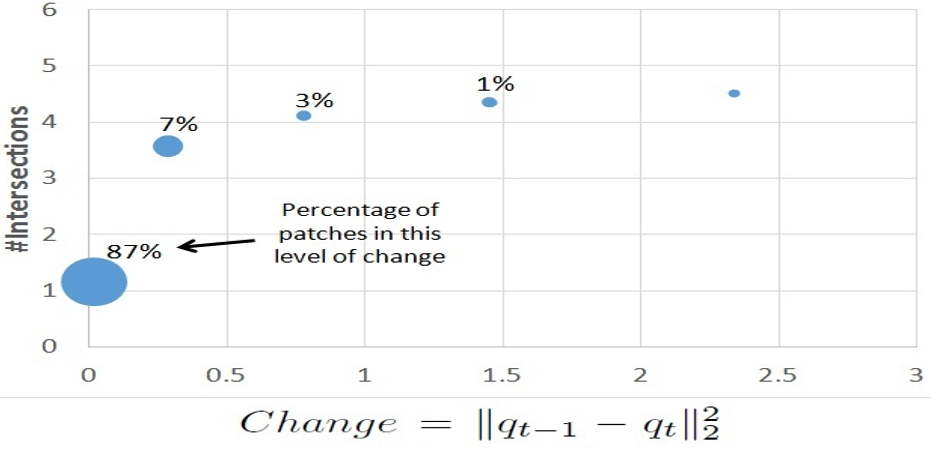}\\
  \includegraphics[width=0.65\linewidth]{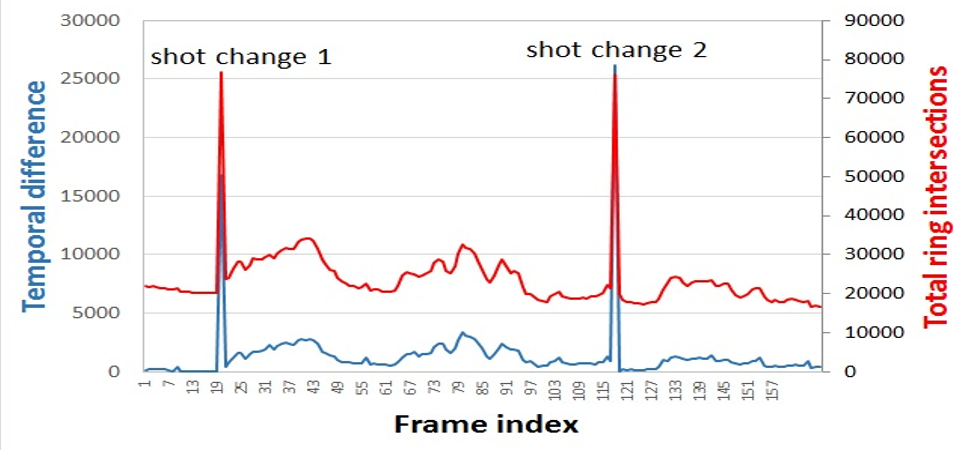}\\
  \end{tabular}
  \caption{Search complexity determined by temporal change: (Top) Patch-level analysis: The mean number of ring intersections as a function of the temporal change $\|q_{t-1}-q_t\|$. (Bottom) Frame-level analysis: The red curve indicates the total number of ring intersections performed over all queries in each frame. The blue curve represents the overall temporal difference between consecutive frames. This shows a clear correlation between the number of rings and the temporal difference between frames.}\label{fig:totIters}
\end{figure}

Figure~\ref{fig:totIters} explores empirically the relation between the number of rings and the amount of temporal change. It shows that the larger the change, the more rings are needed, and the higher the computation time is. As was shown in Figure~\ref{fig:appearanceNgbors}.(b), for most patches the change is very small, and hence the overall computation time is small.

\subsection{RIANN - Our Algorithm}
We name this approach RIANN (Ring Intersection ANN). RIANN is outlined in Algorithm~\ref{Alg:RIANNonline}. Viewing this process as a hashing scheme, we say that RIANN is a query-sensitive hashing since it builds bins around the queries, as opposed to non-sensitive hashing that creates the bins beforehand. Query-sensitive hashing avoids issues where a query lies at a boundary of a bin, thus reducing the chance of better candidates lying in neighbor bins.
Temporal coherency of adjacent frames leads to most queries lying in the vicinity of the current best match. Here, few intersections are often enough to provide very few candidates.


RIANN can find ANNs for each query patch, given the ANN of its predecessor patch and a reference set. Hence, to complete our solution for computing a dense ANN field for video, we need to define two more components: (i) how the reference set is constructed, and (ii) what we do to initialize the first frame.\\

\noindent \textbf{Building a reference model:}
We start by collecting a large set of patches.
To build a {\em global} reference set, that can be used for many target videos, the patches are extracted randomly from a number of natural images.
Since natural images exhibit high redundancy, the collected set of patches is likely to include many similarities.
Inspired by dictionary construction methods such as~\cite{ksvd,regression-tree} we seek a more compact set that represents well these patches.
To dilute patches with high resemblance we cluster the patches using a high dimensional regression tree~\cite{regression-tree}.
We then take the median of each cluster as the cluster representative and normalize it to be of length $1$.
Our final reference set consists of the normalized representatives (query patches are normalized at runtime).
Last, we calculate the pairwise distance matrix and sort it such that column $i$ contains an ascending order of distances from patch $i$.

In some applications one might want to use a reference set that consists only of patches taken from the query video itself.
For example, for methods inspired by Non-Local-Means.
When this is the case, we randomly select a single frame from the video and collect all its patches.
We then again cluster these patches using a high dimensional regression tree~\cite{regression-tree} and take the cluster representatives as our reference set.
We refer to this as a {\em local} reference set.\\

\noindent \textbf{Initialization:}
RIANN requires an initial ANN field of matches for the first frame. We found that initializing these matches randomly, leads to convergence after very few frames. Hence, this was our initialization approach.

\begin{algorithm}
\caption{RIANN: online search}
\label{Alg:RIANNonline}
\begin{algorithmic}[1]
\STATE \textbf{Input:} video $V$, reference patch set $R$, sorted distance matrix $D$ , maximum set size $L=20$, ring width parameter $\alpha=0.25$
\STATE \textbf{Output:} $\{ANNF^{(t)}\}$ - dense ANN field for each frame $t=0,1,2,...$
\STATE \textbf{Initialize:} $ANNF^{(0)}(x,y) \backsim Uniform[1,|R|]$
\STATE \textbf{for t=1,2,...}
\STATE \hspace{.5em} \textbf{for each query patch $q_{x,y,t} \in V^t$:}\
\STATE \hspace{.5em} \hspace{.5em} $q_{x,y,t}\leftarrow \frac{q_{x,y,t}}{\|q_{x,y,t}\|}$
\STATE \hspace{.5em} \hspace{.5em} $r_i = ANNF^{(t-1)}(x,y)$
\STATE \hspace{.5em} \hspace{.5em} $d_i=dist(q_{x,y,t},r_i)$ ; $\varepsilon=\alpha d_i$
\STATE \hspace{.5em} \hspace{.5em} Initial candidate set:
\STATE \hspace{.5em} \hspace{.5em} \hspace{.5em} $S_{x,y,t}=\{r_j \; , \; s.t. \; : \; d_i - \varepsilon \leq dist(r_i,r_j) \leq d_i + \varepsilon\}$
\STATE \hspace{.5em} \hspace{.5em} \textbf{while} $|S_{x,y,t}|\geq L$
\STATE \hspace{.5em} \hspace{.5em} \hspace{.5em} Choose a random anchor point $r_k \in S_{x,y,t}$
\STATE \hspace{.5em} \hspace{.5em} \hspace{.5em} $d_k=dist(q_{x,y,t},r_k)$ ; $\varepsilon=\alpha d_k$
\STATE \hspace{.5em} \hspace{.5em} \hspace{.5em} Update candidate set:
\STATE \hspace{.5em} \hspace{.5em} \hspace{.5em} \hspace{.5em} $S_{x,y,t} = S_{x,y,t} \cap \{r_j \; , \; s.t. \; : \; d_k - \varepsilon \leq dist(r_k,r_j) \leq d_k + \varepsilon\}$
\STATE \hspace{.5em} \hspace{.5em} \textbf{end while}
\STATE \hspace{.5em} \hspace{.5em} Find best match for $q_{x,y,t}$ in $S_{x,y,t}$:
\STATE \hspace{.5em} \hspace{.5em} \hspace{.5em} $ ANNF^{(t)}(x,y) \gets \operatorname{arg\,min}_{s\in S_{x,y,t}}{\{dist(q_{x,y,t},s)\}}$
\end{algorithmic}
\end{algorithm}

\section{Empirical evaluation}
\label{sec:evaluate}
Experiments were performed on a set of $20$ videos from the Hollywood2 dataset ~\cite{hollywood2}. Each video consists of 200 frames and is tested under three common video resolutions: VGA (480x640), SVGA (600x800), and XGA(768x1024). For each video we compute the ANN field for all frames.
We then reconstruct the video by replacing each patch with its match and averaging overlapping patches. To asses the reconstruction quality we compute the average normalized reconstruction error $E=\frac{\|f_{GT} - f_R\|_2}{\|f_{GT}\|_2}$, where $f_{GT}$ is the original frame and $f_R$ is the reconstructed frame by each method. Our time analysis relates to online performance, excluding pre-processing time for all methods.
Patches are of size $8\times8$.

To asses RIANN we compare its performance against previous work.
We do not compare to optical-flow methods, since as discussed in Section~\ref{sec:related-work}, they solve a different problem where matches are computed between pairs of consecutive frames.
One could possibly think of ways for integrating optical flow along the video in order to solve the matching to a single fixed reference set. However, this could require the development of a whole new algorithm.

We do compare to three existing methods for image ANNF: PatchMatch~\cite{PatchMatch}, CSH~\cite{CSH}, and TreeCANN~\cite{TreeCANN}.
These methods require a single image as a reference, therefore, we take for each video a single random frame as reference.
For RIANN, we experiment with both a {\em local} model built from the same selected frame, and with a {\em global} reference model built from patches taken from multiple videos.
The same global reference model was used for all $20$ target videos.
We do not compare to the video-ANNF of~\cite{liu2010highquality} since they report executing 4 iterations of PatchMatch-like spatial propagation in addition to optical flow computation. Their solution is thus slower than PatchMatch, which is the slowest of the methods we compare to.

\begin{figure*}[t]
  \centering
  \begin{tabular}{cc}
    \includegraphics[width=0.44\textwidth]{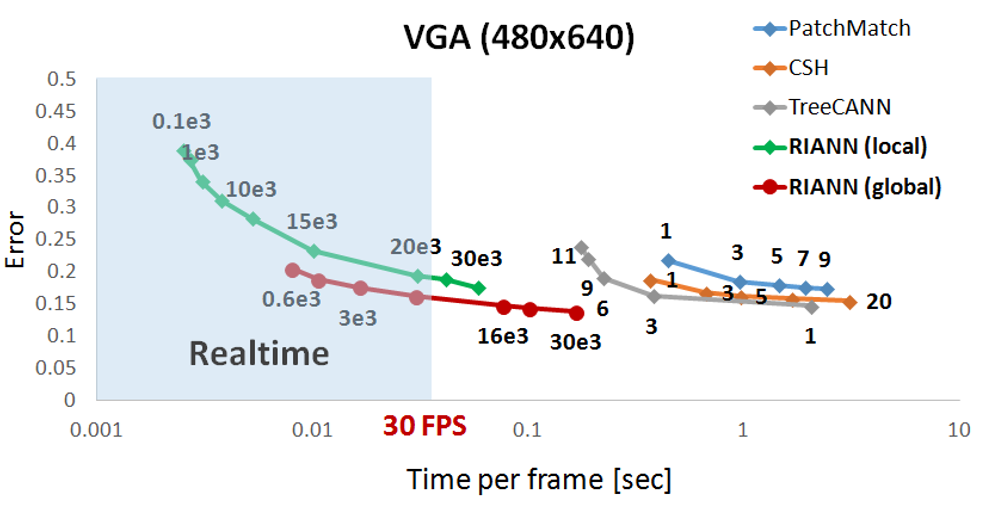}&
    \includegraphics[width=0.44\textwidth]{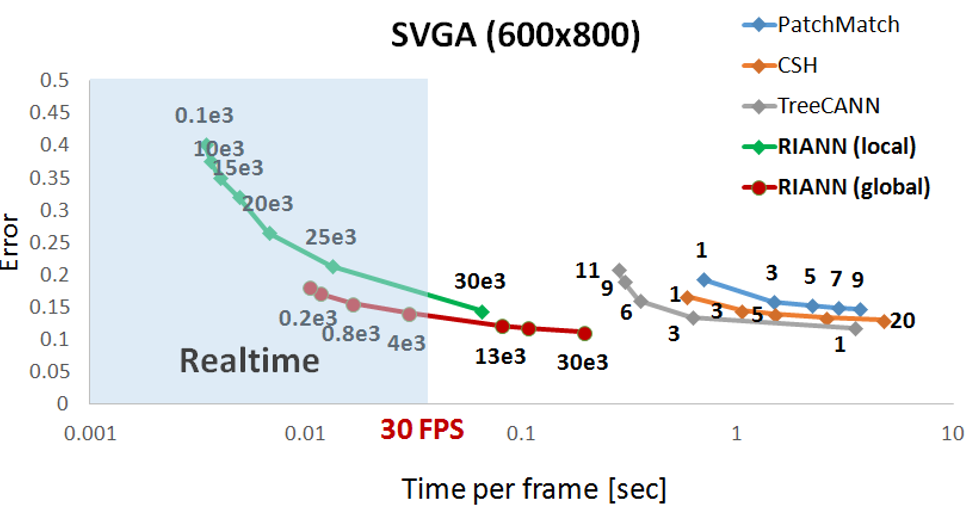}\\
    (a) & (b)\\
    \includegraphics[width=0.44\textwidth]{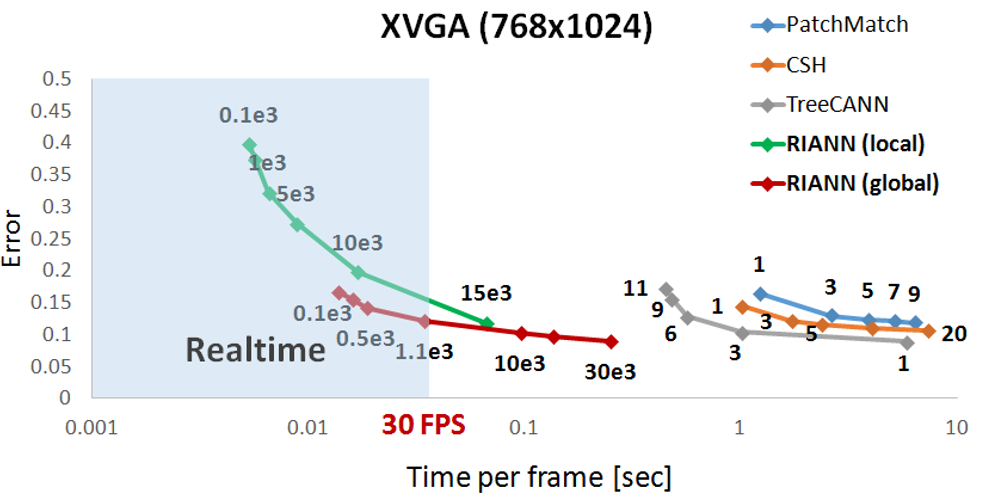}&
    \includegraphics[width=0.44\textwidth]{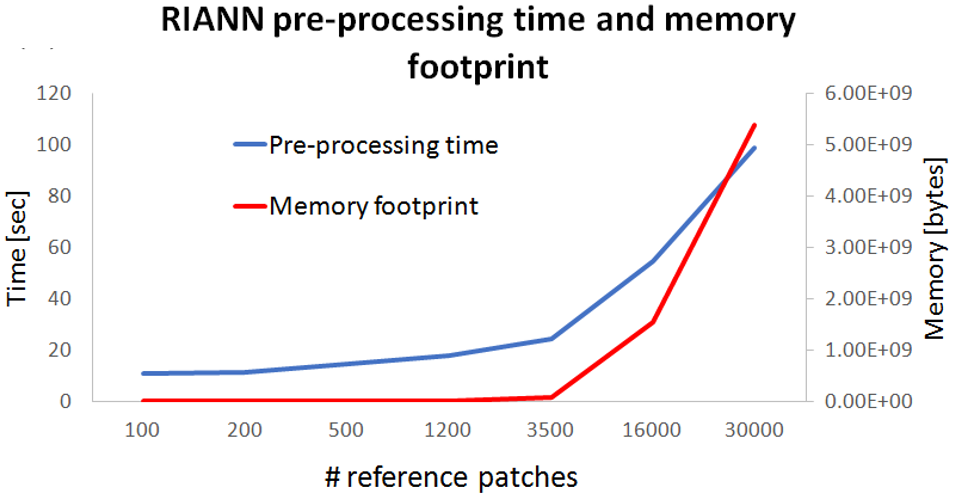}\\
    (c)&(d)
  \end{tabular}
 \caption{Runtime vs. accuracy tradeoffs. (a-c) A comparison of runtime and accuracy shows RIANN is significantly faster than previous approaches. In fact, RIANN is the only realtime solution. The curves for PatchMatch and CSH represent results for varying numbers of iterations. For TreeCANN we vary the grid sparsity, and for RIANN the reference set size. The labels next to the tickmarks correspond to the iterations (PatchMatch,CSH), grid-size (TreeCANN), and set-size (RIANN), correspondingly.
Pre-processing time was excluded for all methods. For RIANN-global a single pre-processed reference set was used for all the test videos. For CSH, TreeCANN and RIANN-local a single frame was pre-processed for each video. PatchMatch has no pre-processing.  
(d) The curves present the pre-processing time (blue) and memory footprint (red) of RIANN, for reference sets of corresponding sizes.}\label{image-results}
\end{figure*}

Figure~\ref{image-results} displays the results. As can be seen, RIANN is the only patch-matching method that works in realtime. It offers the widest tradeoff between speed and accuracy, making it a good option for a wide range of applications. RIANN's fastest configurations are up to hundreds of frames per second, making it attractive to be used as a preliminary stage in various realtime systems. In addition, RIANN offers configurations that outperform previous state of the art algorithms~\cite{PatchMatch,CSH,TreeCANN} in both accuracy and speed.

In the experiments presented in Figure~\ref{image-results} the ANNF for each frame was computed independently of all other frames. 
A possible avenue for improving the results of image-based methods is to initialize the matches for each frame using the matches of the previous frame.
We have tried this with PatchMatch and found that the speed barely changed, while accuracy was slightly improved. 
For CSH and even more so for TreeCANN doing this is not straightforward at all, e.g., for TreeCANN one would need a new search policy that scans the tree from the leaves (the solution of the previous frame) rather than the root. 
We leave this for future work.

Changing the size of the reference set affects our time-accuracy tradeoff. A large reference set helps approximate the queries with greater precision on the account of a slower search. A small reference set leads to a quicker search at the account of lower accuracy.
It can be seen that when we use a {\em local} reference-set, i.e., when the patches are taken from a single frame, RIANN's accuracy is lower than that of the other methods. This is probably since we cluster the patches while they use the raw set. However, this is resolved when using a {\em global} reference set constructed from patches of several different videos. In that case RIANN's accuracy matches that of previous methods.

A disadvantage of RIANN is a bad scaling of memory footprint ($O(n^2)$), resulting from the storage of the $n\times n$ matrix $D$ of distances between all reference patches ($n$ is the reference set size). The pre-processing time ranges from $\sim10$[sec] for small models up to $\sim2$[min] for the maximum size tested, see Figure~\ref{image-results}.(d).
Hence, using a single global model is advantageous.

\section{RIANN for Real-Time Video Processing}
\label{sec:ann-based-video-proessing}

\begin{figure*}[t]
  \centering
\begin{tabular}{cccccc}
\rotatebox{90}{\ \ \ Input: gray}&
 \includegraphics[width=0.15\textwidth]{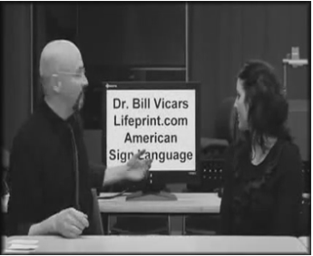}&
 \includegraphics[width=0.15\textwidth]{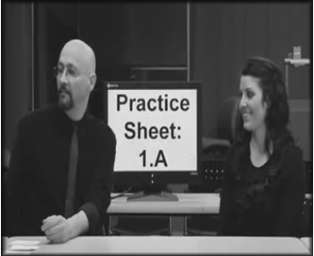}&
 \includegraphics[width=0.15\textwidth]{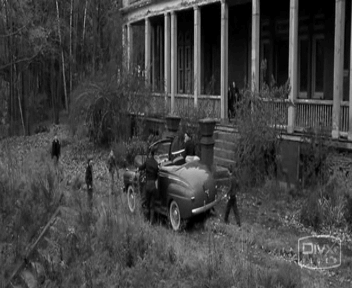}&
 \includegraphics[width=0.15\textwidth]{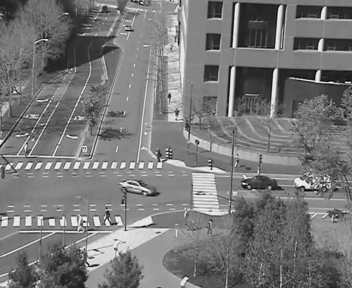}&
  \includegraphics[width=0.15\textwidth]{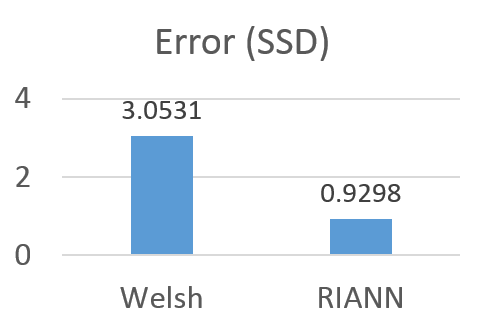}\\
\rotatebox{90}{\ \ \ Output: color}&
 \includegraphics[width=0.15\textwidth]{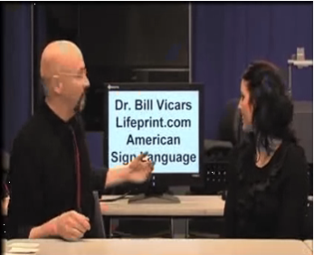}&
 \includegraphics[width=0.15\textwidth]{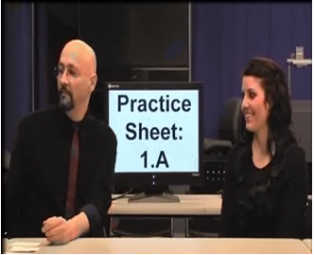}&
 \includegraphics[width=0.15\textwidth]{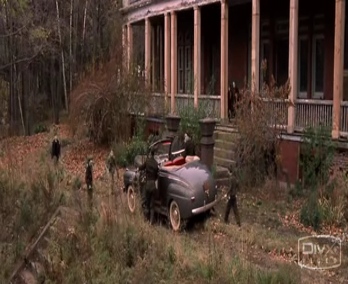}&
 \includegraphics[width=0.15\textwidth]{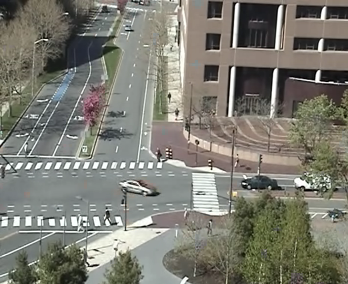}&
  \includegraphics[width=0.15\textwidth]{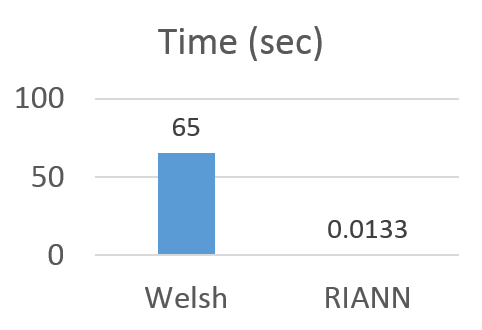}\\
\end{tabular}
 \caption{Colorization: (Top) Example frames from input grayscale videos. (Bottom) The corresponding colored versions, computed automatically, at 30FPS, using RIANN. The colorization is quite sensible, with occasional errors in homogenous regions.
 (Right) The charts compare run-time (per frame) and Sum of Squared Differences (per pixel) averaged over our test set.
 RIANN is both faster and more accurate than Welsh et al.~\cite{welsh2002transferring}}
 \label{fig:colorization}
\end{figure*}

The proposed ANN framework could be useful for several applications. For example, in video conferencing, one could transmit only the grayscale channel and recolor the frames, in realtime, at the recipient end. Alternatively, in man-machine interfaces such as Kinect, one could apply realtime denoising or sharpening to the video. In both scenarios, a short pre-processing stage is acceptable, while realtime performance at runtime is compulsory.

A broad range of effects and transformations can be approximated by video ANN, e.g., cartooning, oil-painting, denoising or colorization. This requires applying the effect/transformation to each patch in the reference set at pre-processing time. Then, at runtime each query patch is replaced with the transformed version of its match in the reference set. Such an approximate, yet realtime, solution could be highly useful for real-world setups, when the complexity of existing methods is high. This is currently the case for several applications. For example, denoising a VGA frame via BM3D \cite{BM3D} runs at $\sim$5[sec], colorization \cite{Colorization} takes $\sim$15[sec], and some of the styling effects in Adobe$^{\circledR}$ Photoshop$^{\circledR}$ are far from running at realtime.

For the approximation to be valid, the applied transformation $T$ should be Lipschitz continuous at patch level: $dist(T(q),T(r)) \leq \alpha \cdot dist(q,r)$, where $\alpha$ is the Lipschitz constant, $q$ is a query patch and $r$ is its ANN.
$\alpha$-Lipschitz continuous process guarantees that when replacing a query patch $q$ with its ANN $r$, then $T(r)$ lies in an $\alpha \cdot dist(q,r)$-radius Ball around $T(q)$. A smaller $\alpha$ implies that $T$ is approximated more accurately.

Lipschitz continuity of image transformations depends on the patch size. Working with bigger patches, increases the probability of a transformation $T$ to be $\alpha$-Lipschitz continuous. For larger patches typically the matches are less accurate and the approximation error is larger. Therefore, to maintain low errors one needs a larger reference set as the patch size increases. To see this consider the extreme case where the patch size is $1\times1$. In this case, using a very small set of 255 patches (one for each gray level), a perfect reconstruction of any gray-level video is possible. However, most transformations are far from being Lipschitz continuous at this scale, hence they cannot be properly approximated via patch-matching. Working with bigger patches increases the probability for Lipschitz continuity, but at the same time demands a larger reference set.

Many video scenarios are characterized by high redundancy, e.g., video chatting, surveillance, indoor scenes, etc. In these scenarios a small reference set suffices to get satisfactory approximation quality. Our test set includes generic videos from the Hollywood2 dataset. Nevertheless, RIANN provided satisfactory results for these videos, for a variety of applications (frame resolution was $288\times352$, patch size $8\times8$). Some example results of edited videos are provided in the supplementary. We next explain how some of these effects were realized and compare to prior art.

\noindent \textbf{Realtime Video Colorization:}
To color a grayscale video we construct a reference set of grayscale patches for which we have available also the color version.
We use RIANN to produce the ANN Fields in realtime.
Each query grayscale patch is converted to color by taking the two chromatic channels of the colored version of its ANN.
Since patches of different colors could correspond to the same grayscale patch, the usage of a global reference set could result in inappropriate colors.
Therefore, we use a local reference set constructed from one random frame of the target video.
We generate color versions for its patches by coloring the frame offline manually (or using a semi-automatic coloring method such as~\cite{Colorization}).
The rest of the video is then colored automatically by RIANN.

Example results are displayed in Figure~\ref{fig:colorization}, and in the supplementary.
It can be seen that our results are plausible most of the time, with a few glitches, mostly in homogeneous regions.
We further compare our results with those of Welsh et al.~\cite{welsh2002transferring}, which also provides fully automatic colorization.
Videos of our test set were converted to gray-scale and then re-colored by both RIANN and~\cite{welsh2002transferring}. 
We compare the average $L_2$ distance (SSD) per pixel between the original video and the colored one as well as the run-time. 
The results reported in Figure~\ref{fig:colorization} show that RIANN is 3 orders of magnitude faster and still more accurate than~\cite{welsh2002transferring}. 

\begin{figure*}[t]
  \centering
\begin{tabular}{ccccc}
Noisy Input & Gaussian & BM3D & RIANN \\
 \includegraphics[width=0.15\textwidth]{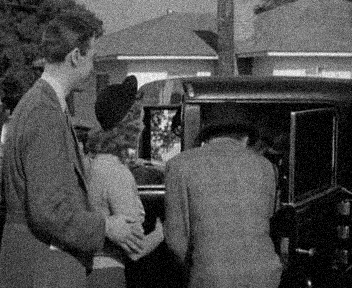}&
 \includegraphics[width=0.15\textwidth]{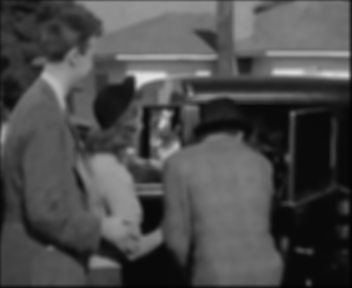}&
 \includegraphics[width=0.15\textwidth]{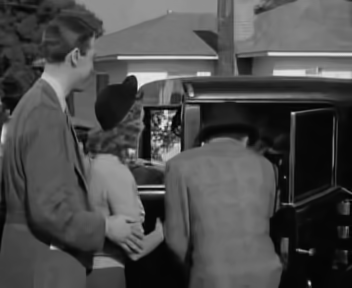}&
 \includegraphics[width=0.15\textwidth]{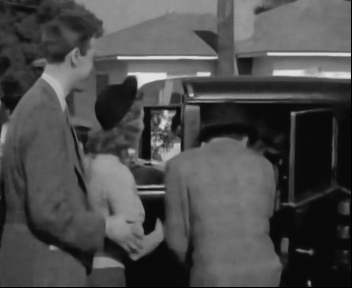}&
 \includegraphics[width=0.2\textwidth]{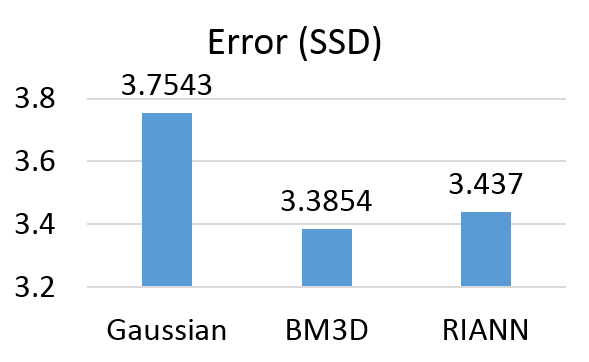}\\
 \includegraphics[width=0.15\textwidth]{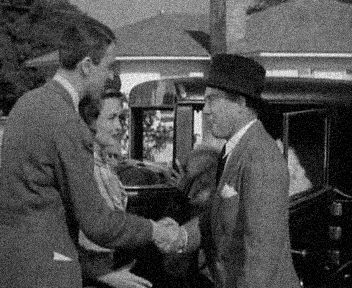}&
 \includegraphics[width=0.15\textwidth]{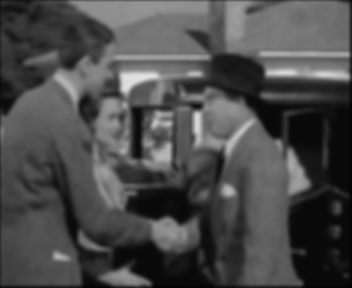}&
 \includegraphics[width=0.15\textwidth]{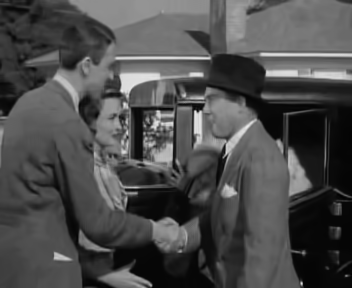}&
 \includegraphics[width=0.15\textwidth]{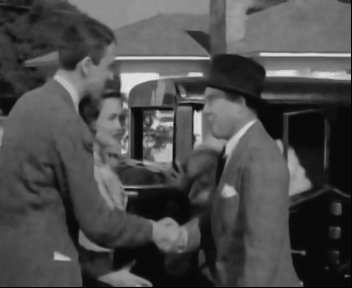}&
 \includegraphics[width=0.2\textwidth]{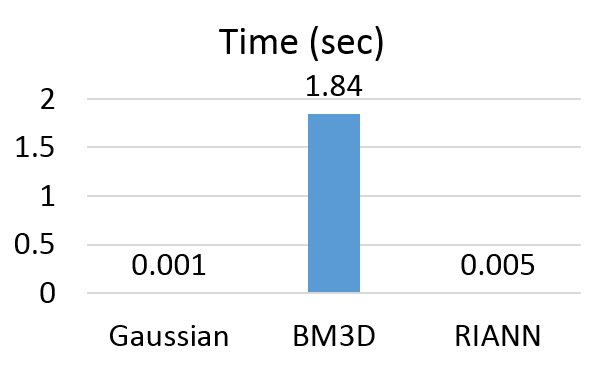}\\
\end{tabular}
 \caption{Denoising: The images are of two example noisy frames and their denoised versions obtained by Gaussian filter, BM3D, and RIANN.
The charts compare run-time (per frame) and Sum of Squared Differences (per pixel) averaged over our test set. 
RIANN is almost as fast as Gaussian filtering while being only slightly less accurate than BM3D.
 }\label{image-denoising}
\end{figure*}

\noindent \textbf{Realtime Video Denoising:}
To test denoising we add Gaussian noise at a ratio of $\frac{\sigma_{noise}}{\sigma_{signal}}=7\%$ . 
We then extract one frame at random, and denoise it using BM3D~\cite{BM3D}. This is used to construct the local reference set. The rest of the frames are denoised online by replacing each patch with the denoised version of its ANN. Example results and quantitative comparisons to BM3D~\cite{BM3D} (accurate) and Gaussian Filter (fast) are provided in Figure~\ref{image-denoising}.
For both BM3D and the Gaussian filter we tuned the parameters to get minimal error.
The results show that our goal was achieved, and while maintaining realtime performance we harm the accuracy only a little.

\begin{figure*}[t]
  \centering
\begin{tabular}{ccccccc}
\rotatebox{90}{\ \ \ Reference}&
\includegraphics[width=0.15\textwidth]{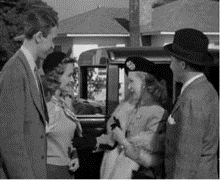}&
\rotatebox{90}{\ \ \ Photoshop}&
\includegraphics[width=0.15\textwidth]{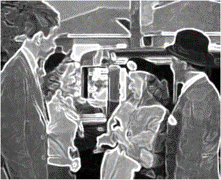}&
\includegraphics[width=0.15\textwidth]{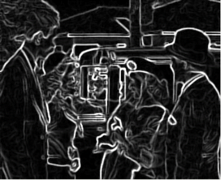}&
\includegraphics[width=0.15\textwidth]{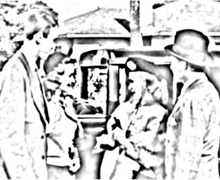}&
\includegraphics[width=0.15\textwidth]{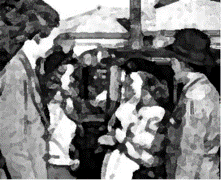}\\
\rotatebox{90}{\ \ \ \ \ Target}&
\includegraphics[width=0.15\textwidth]{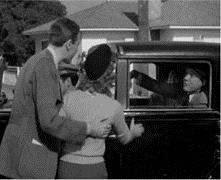}&
\rotatebox{90}{\ \ \ RIANN}&
\includegraphics[width=0.15\textwidth]{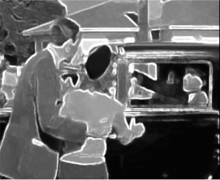}&
\includegraphics[width=0.15\textwidth]{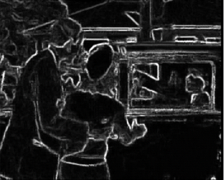}&
\includegraphics[width=0.15\textwidth]{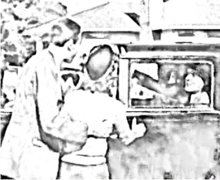}&
\includegraphics[width=0.15\textwidth]{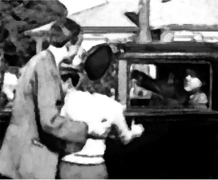}\\
& & & Accent Edges & Glowing Edges & Photocopy & Fresco\\
\end{tabular}
 \caption{Styling effects: (Top) A reference image and several styling effects applied to it in PhotoShop. (Bottom) An example target frame and the approximated styling effects obtained by RIANN at 30 FPS. 
 }\label{fig:photoshop}
\end{figure*}

\noindent \textbf{Realtime Styling Effects:}
To show applicability to a wide range of image transformations, we apply a set of Adobe$^{\circledR}$ Photoshop$^{\circledR}$ effects to one frame of the video. The patches of this frame are used to construct the reference set. We then use RIANN to find ANNs and replace each patch with the transformed version of its match. We tested several effects and present in Figure~\ref{fig:photoshop} sample results for ``Accent Edges'', ``Glowing Edges'', ``Photocopy'' and ``Fresco''.

\section{Discussion on Spatial Coherency in Video}
\label{sec:coherency-spatial}
Spatial coherency is typically used for ANN Fields in images by propagating matches across neighboring patches in the image plane~\cite{PatchMatch}.
At first we thought that spatial coherency would also be useful for video ANN.
However, our attempts to incorporate spatial coherency suggested otherwise. 
Therefore, we performed the following experiment. 
Our goal was to compare the accuracy of matches found when relying on spatial coherency, to the accuracy of matches based on appearance only. To compute the latter, we found for each patch in a given target video, its exact NN in a given reference set. To compute the former, we applied PatchMatch~\cite{PatchMatch} (set for 3 iterations) to the same target video, since PatchMatch relies heavily on spatial coherency. As a reference set we took all the patches of a single random frame of the video.

\begin{figure}[htb]
  \centering
  \includegraphics[width=0.75\linewidth]{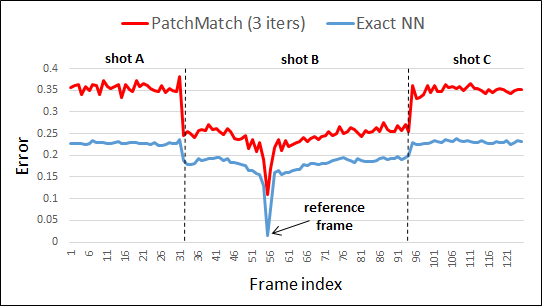}\\
  \caption{Limitations of spatial propagation in video: The curves correspond to the reconstruction error when replacing each patch of a video with its match in a reference frame. PatchMatch \cite{PatchMatch} uses spatial propagation, while exact NN is based solely on appearance. An abrupt increase of PatchMatch error with respect to the exact NN is evident in shots A,C. There, PatchMatch will require more than 3 iterations to achieve a low reconstruction error as in shot B. This happens because spatial propagation is less effective as the dissimilarity between reference and target frames increases.}\label{image-sensitivity2structure}
\end{figure}

Results are presented in Figure~\ref{image-sensitivity2structure}. As can be seen, PatchMatch error jumps significantly when the target and reference frames are from different shots. This occurs since there is low spatial coherency across visually different frames. On the contrary, the error of appearance-based matching increases only slightly across shot changes.
This supports our approach, that relies solely on temporal coherency to propagate neighbors in appearance space.

\section{Conclusion}
\label{sec:conclusion}
We introduced RIANN, an algorithm for computing ANN Fields in video in realtime. It can work with any distance function between patches, and can find several closest matches rather than just one. These characteristics could make it relevant to a wider range of applications, such as tracking or object detection.
RIANN is based on a novel hashing approach: query-sensitive hashing with query-centered bins. This approach guarantees that queries are compared against their most relevant candidates. It could be interesting to see how these ideas translate to other hashing problem.


\paragraph{Acknowledgements:}
This research was funded (in part) by Minerva, the
Ollendorf Foundation, and the Israel Science Foundation under Grant 1179/11.

{\small
\bibliographystyle{ieee}
\bibliography{egbib}
}
\end{document}